\def\year{2020}\relax
\documentclass[letterpaper]{article} 
\usepackage{aaai20}  
\usepackage{times}  
\usepackage{helvet} 
\usepackage{courier}  
\usepackage[hyphens]{url}  
\usepackage{graphicx} 
\usepackage{graphicx}  

\usepackage{amsmath}
\usepackage{amssymb}
\usepackage{subcaption}
\usepackage{booktabs, makecell, tabularx}
\usepackage{epstopdf}
\usepackage{enumitem}
\usepackage{mathtools}
\usepackage{pifont}
\usepackage{microtype}

\nocopyright
\title{Geometry-constrained Car Recognition Using a 3D Perspective Network}
\author{Rui Zeng\textsuperscript{\rm 1\rm2}, Zongyuan Ge\textsuperscript{\rm 2$\ast$}, Simon Denman\textsuperscript{\rm 1}, Sridha Sridharan\textsuperscript{\rm 1}, Clinton Fookes\textsuperscript{\rm 1} \\ 
\textsuperscript{\rm 1} Queensland University of Technology \hspace{2cm} \textsuperscript{\rm 2} Monash University \\
{r3.zeng}@hdr.qut.edu.au; {\{s.denman, s.sridharan, c.fookes\}}@qut.edu.au; zongyuan.ge@monash.edu
}

\begin{document}

\maketitle

\begin{abstract}
We present a novel learning framework for vehicle recognition from a single RGB image. Unlike existing methods which only use attention mechanisms to locate 2D discriminative information, our work learns a novel 3D perspective feature representation of a vehicle, which is then fused with 2D appearance feature to predict the category. The framework is composed of a global network (GN), a 3D perspective network (3DPN), and a fusion network. The GN is used to locate the region of interest (RoI) and generate the 2D global feature. With the assistance of the RoI, the 3DPN estimates the 3D bounding box under the guidance of the proposed vanishing point loss, which provides a perspective geometry constraint. Then the proposed 3D representation is generated by eliminating the viewpoint variance of the 3D bounding box using perspective transformation. Finally, the 3D and 2D feature are fused to predict the category of the vehicle. We present qualitative and quantitative results on the vehicle classification and verification tasks in the BoxCars dataset. The results demonstrate that, by learning such a concise 3D representation, we can achieve superior performance to methods that only use 2D information while retain 3D meaningful information without the challenge of requiring a 3D CAD model.
\end{abstract}
\section{Introduction}
\begin{figure}[t]
	\centering
	\includegraphics[width=1\linewidth]{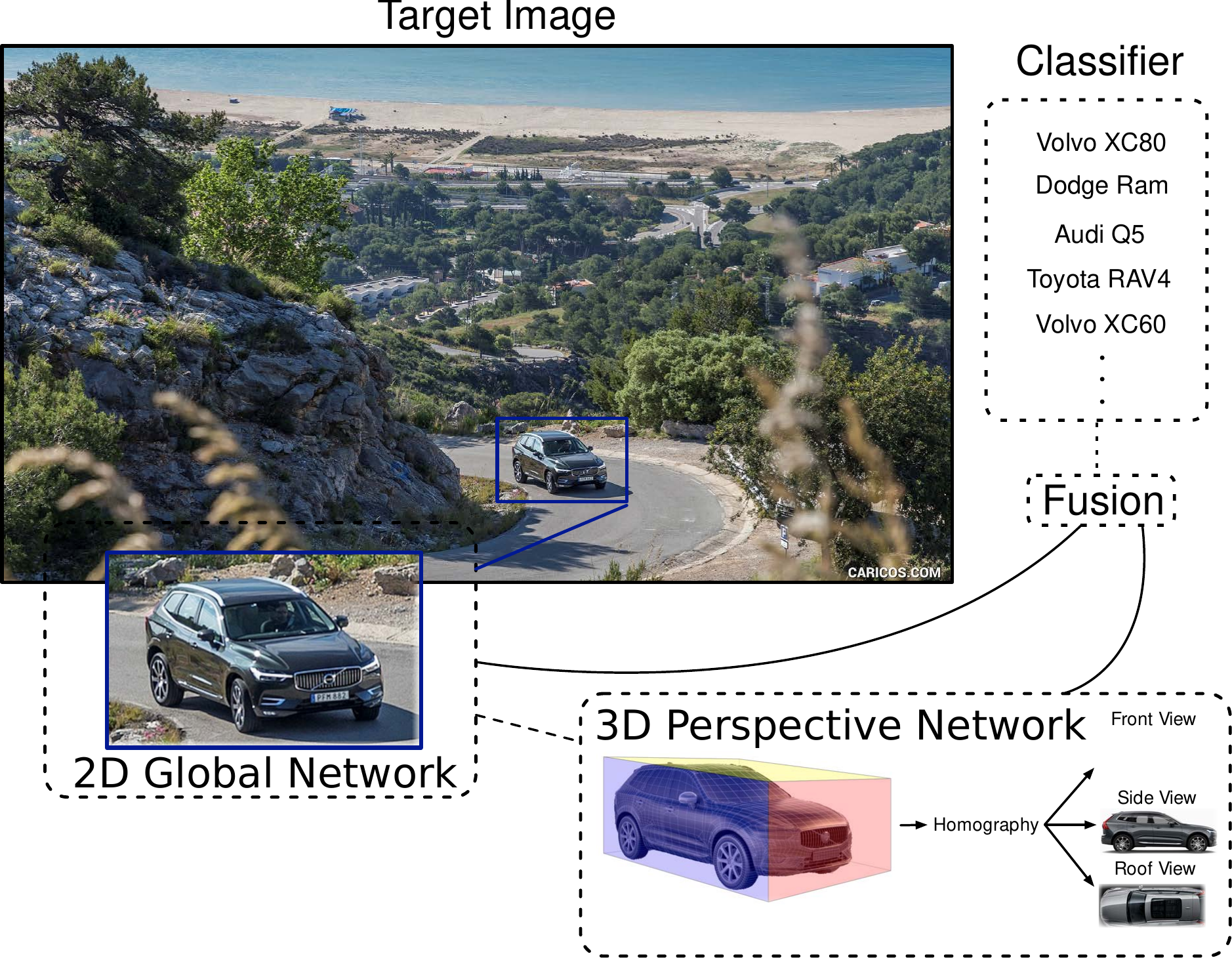}
	\caption{Consider the image of the car shown above. Even though the vehicle is shown on a flat 2D image, our model can estimate and leverage knowledge from 2D appearance as well as the rigid 3D bounding box of the vehicle to produce a viewpoint-normalized representation, which is able to improve vehicle recognition performance.}
	\label{fig:3dconcept}
\end{figure}
\noindent
Traffic surveillance systems are an important part of intelligent transportation,\let\thefootnote\relax\footnote{$\ast$ indicates the corresponding author.} which is the core of artificial intelligence (AI) in smart cities. A holy grail for traffic surveillance is the ability to automatically recognize and identify vehicles from visual information alone. Vehicle recognition enables automated car model analysis, which is helpful for innumerable purposes including regulation, description, and indexing vehicles.

One key idea shared by recent vehicle recognition algorithms is to use an ensemble of local features extracted from discriminative parts of the vehicle, which can be located using either part annotations or attention mechanisms. These approaches, given part annotations, \cite{krause2014learning,he2015recognition} learn the corresponding part detectors and then assemble these to obtain a uniform representation of the vehicle, which is used for category classification. To overcome the need for part annotations, recent advances \cite{jaderberg2015spatial,yang2018learning,wang2018learning,fu2017look} make use of attention mechanisms to identify salient spatial regions automatically. Despite these part-aware methods successfully leveraging spatial information, they are still `flat', i.e., built on independent and 2D views.

3D-aware methods have been shown to be promising alternatives to part-aware approaches. For instance, \cite{krause20133d,lin2014jointly} exploit the fact that aligning a 3D CAD model or shape to 2D images significantly eliminates the variation caused by viewpoint changes, which is shown as the main obstacle for vehicle categorization. However, these methods have limited generality as they require 3D CAD models for vehicles.

To address these issues, we instead propose a concise 3D representation for vehicle recognition by directly using the 3D bounding box. Our work is summarized in Figure~\ref{fig:3dconcept}. Our method has three components: the Global Network (GN), the 3D Perspective Network (3DPN), and the Feature Fusion Network (FFN). The GN detects and extracts relevant global appearance features of vehicles from input RGB images. The 3DPN predicts the 3D bounding box under the geometric constraints of the vanishing points using the proposed vanishing point loss. With the assistance of the predicted 3D bounding box, the 3DPN further generates a viewpoint-aligned feature representation in a geometrically correct manner. Finally, the features generated from the GN and the 3DPN are merged in the FFN and then used for vehicle recognition. Our contributions can be summarized as follows:
\begin{itemize}[leftmargin=*,align=left,itemsep=2pt,parsep=0pt]
	\item We propose a concise 3D representation, which is termed as 3D perspective feature, for vehicle recognition. The proposed representation uses 3D information in a meaningful and correct manner without the challenge of requiring a 3D CAD model. Based on the proposed method, a unified network architecture for vehicle recognition which takes full advantage of the 2D and 3D representations is presented.
	\item We introduce a geometrically interpretable loss (vanishing point loss) to elegantly enforce the consistency of the predicted 3D bounding box to improve regression accuracy.
	\item We evaluate our proposed method on the vehicle classification and verification tasks in the BoxCars benchmark and achieve the state-of-the-art results.
\end{itemize}
\section{Related Work}
\noindent
We review the previous works on vehicle recognition and 3D bounding box estimation, which are related to our approach.
\subsection{Vehicle Classification}
Since our model uses only a single image to recognize vehicles, methods which use extra information, such as 3D CAD models, are not reviewed. 2D vehicle recognition can be classified into two categories: part-annotation (PA) and attention-mechanism (AM) methods.

While PA methods \cite{krause2014learning,he2015recognition,sochor2016boxcars} are able to achieve high performance by extracting local feature representation from detected vehicle parts, they are reliant on part annotations. The labor intensive annotation is usually not possible during inference when applying such methods to a real scene. \cite{he2015recognition} detects each discriminative part of a vehicle and then generates a uniform feature using the HOG descriptor. \cite{krause2014learning} trains a classification CNN by combining both local and global cues, which have been previously annotated. Similarly, \cite{sochor2016boxcars} uses a pre-annotated 3D bounding box to generate a 2D ``flat" representation.

To alleviate the essential requirement of annotations, AM methods \cite{yang2018learning,fu2017look,jaderberg2015spatial,wang2018learning} have been extensively researched in recent years. One common feature of them is to locate discriminative parts of a vehicle automatically using attention mechanisms. \cite{jaderberg2015spatial} aims to determine an affine transformation to map a entire vehicle to its most discriminate viewpoint in a global way. \cite{yang2018learning,fu2017look,wang2018learning} generate discriminative features locally by searching salient primitives, and then use them for recognition.

In contrast to previous methods, we take a further step towards taking full advantages of both the 2D and 3D representation of a vehicle. Comparing with PA and AM methods, our method is able to predict the 2D and 3D bounding box simultaneously. It can generate viewpoint normalization features using appropriate geometric constraints in a geometrically explainable way. While \cite{manhardt2019roi,simonelli2019disentangling} both leverage the 3D box for feature guidance, our method generates features from a 3D box using perspective transformations, which enhances recognition performance. Moreover, compared to 3D-aware methods, our method is totally free from 3D CAD models, which are difficult to obtain in practice.
\subsection{3D Bounding Box Estimation}
Vehicle 3D bounding box estimation has been an active research topic in recent years. \cite{mousavian20173d} regresses vehicles' dimensions and constructs 3D bounding boxes using camera intrinsic parameters. \cite{xu2018pointfusion} estimates the 3D bounding box of a vehicle using the combination feature of the point cloud and the 2D image. Our goal is to predict the 3D bounding box of a vehicle only using the RGB image. Therefore we seek to predict eight vertices directly. \cite{hedau2012recovering,gupta20113d} localize vertices using corner detectors and then construct 3D bounding box through the geometric relationships among all vertices. Following on the success of these geometry-based methods, DeepCuboid \cite{dwibedi2016deep} regresses vertices of the 3D bounding box through a Faster-RCNN-based model. Subsequently, vertex predictions are refined by utilizing vanishing points \cite{hartley2003multiple}. However, this refinement step is separate from the network training stage, and the vanishing points computed from inaccurate predictions often lead to significant error.

Unlike \cite{dwibedi2016deep}, we use the proposed vanishing point (VP) regularization to encode the VP constraint of the eight vertices during network training. It allows our model to avoid any post refinement to redress vertices.
\section{Methodology}
\label{sec:fgcn}
\begin{figure*}[t]
	\centering
	\includegraphics[width=1\linewidth]{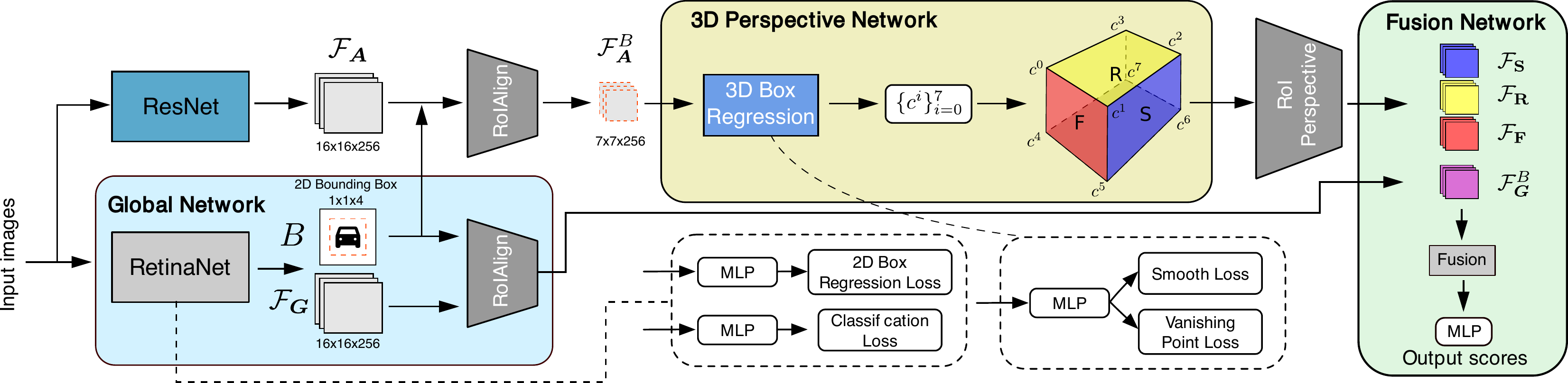}
	\caption{Overview of the proposed model. The model is composed of three main components: (A) Global Network (GN), which aims to localize the vehicle and extract its 2D features. (B) 3D Perspective Network (3DPN), which performs 3D bounding box regression by taking the anchor from the predicted 2D bounding box and generates 3D perspective features of the three main faces (front, roof, and side) of the vehicles. (C) Feature Fusion Network, which fuses the features from the \textbf{GN} and \textbf{3DPN} by applying multi-modal compact bilinear (MCB)~\cite{fukui2016multimodal} pooling. {\textbf{F}}, {\textbf{R}}, {\textbf{S}} in the predicted 3D bounding box represents the front/rear, roof, and side respectively.}
	\label{fig:detailed_art_new}
\end{figure*}
\subsection{Overview}
Our goal is to design an architecture that jointly extracts features in terms of both the 2D and 3D representation for vehicle recognition. Figure \ref{fig:detailed_art_new} shows the overview framework of the proposed method.
\subsection{Global Network (GN)}
\label{sec:gfn}
\noindent
The GN uses a variant of RetinaNet \cite{lin2018focal} to localize the vehicle using a 2D bounding box. RetinaNet is a dense object detector composed of a CNN-FPN \cite{lin2017feature} backbone, which aims to extract a convolutional feature map, $\mathcal{F}_{\boldsymbol{G}}$, over an entire input image.
Two task-specific subnetworks are attached to the backbone network to perform object classification and 2D object box regression respectively. RetinaNet offers comparable performance to complex two-stage detectors~\cite{ren2015faster,he2017mask} while retaining the advantages of one-stage detectors such as inference speed and model optimization. These attributes are desirable when adopting it as one component in an end-to-end classification framework. To adapt the original RetinaNet as the part of our network, we make the following modifications:
\subsubsection{ROIAlign}
We add an ROIAlign layer~\cite{he2017mask} after the 2D box decoding process.
The detected 2D bounding box of the vehicle is denoted $B=(B_x, B_y, B_w, B_h)$, where $B_x$, $B_y$ are the left-top corner coordinates with respect to $x$ and $y$ axis. $ B_w$, $B_h$ represents the width and height of $B$ respectively.
The ROIAlign layer combined with the detected 2D bounding box coordinates is able to produce a fixed-sized global feature representation which comprises the vehicle, termed $\mathcal{F}^{B}_{\boldsymbol{G}}$.
In particular, this modification ensures that errors in the extracted 2D coordinates can be back propagated through the GN when trained jointly with other network components.
\subsection{3D Perspective Network (3DPN)}
\label{sec:3D}
\noindent
Figure \ref{fig:detailed_art_new} illustrates the architecture of the 3DPN.
Its role is to provide geometrically-interpretable features by normalizing the vehicle viewpoint to account for perspective distortion. To achieve this, the 3DPN takes as input $\mathcal{F}_{\boldsymbol{A}}^{B}$, which is the feature map pooled from $B$ at $\mathcal{F}_{\boldsymbol{A}}$ using RoIAlign. $\mathcal{F}_{\boldsymbol{A}}$ is the auxiliary feature map extracted from an off-the-shelf CNN. We then estimate the coordinates of eight vertices' of the 3D bounding box, $C: \{c^i\}_{i=0}^{7}$, using a 3D bounding box regression network. Subsequently, $C$ is used to generate 3D perspective feature using perspective transformation in feature-map level.  normalize the viewpoint of the vehicle in $\mathcal{F}_{\boldsymbol{A}}^{B}$ using perspective transformation. As a result, $\mathcal{F}_{\textbf{R}}$, $\mathcal{F}_{\textbf{F}}$, and $\mathcal{F}_{\textbf{S}}$, representing perspective transformed feature maps from the quadrilaterals formed by the roof (\textbf{R}), front (\textbf{F}), and side (\textbf{S}) of the vehicle, are extracted. Below we describe the 3D bounding box regression network with the proposed vanishing point loss, and 3D perspective feature respectively. 
\subsubsection{3D bounding box regression branch}
Instead of using the absolute coordinates of the 3D box in the image space directly, we estimate them in an RoI relative coordinate system by leveraging the 2D bounding box as an anchor. For each $\{c^i\}_{i=0}^{7}$ in the image coordinate system we first transform those points to the 2D-bounding-box relative coordinate system: $\hat{c}^i_x = (c^i_x - B_x - B_w/2 )/ B_w$, and $\hat{c}^i_y = (c^i_y - B_y - B_h/2 )/ B_h$, where $\{\hat{c}^i\}_{i=0}^7$ is the training target of this branch. The 3D bounding box regression network takes $\mathcal{F}_{\boldsymbol{A}}^B$ as the input feature map. Then it applies two conv layers ($3\times3\times256$) and a multilayer perceptron ($512\times16$) to regress all $x$ and $y$ coordinates of $\{\hat{c}^i\}_{i=0}^7$ (leaky ReLu are used as activations). The loss function used to train this sub-network is:
\begin{equation}
L_{\text{3Dbranch}}= {L_{\text{smooth}l_1}(\hat{c}^\ast, \hat{c}) + L_{\text{vp}}},
\label{eqn:bb3d}
\end{equation}
where $\hat{c}^\ast$ is the ground-truth locations for $\hat{c}$, $L_{\text{smooth}l_1}$ is the standard smooth-$l_1$ loss and $L_{\text{vp}}$ is the proposed vanishing point regularization loss to ensure that $C$ satisfies perspective geometry (i.e., every parallel edge of $C$ intersects at the same vanishing point).
\subsubsection{Vanishing point regularization}
\begin{figure}[t]
	\includegraphics[width=1\linewidth]{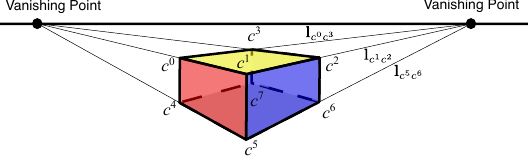}
	\caption{Illustration of the vanishing point. To simplify the visualization, only two vanishing points (in the {\textbf{F}} and {\textbf{S}} directions) are plotted. The lines ${\textbf{l}}_{c^0c^3}$, ${\textbf{l}}_{c^1c^2}$, and ${\textbf{l}}_{c^5c^6}$ contribute to the first part of $L_{\text{vp}_{\textbf{F}}}$.}
	\label{fig:vploss}
\end{figure}

A standard smooth-$l_1$ loss lacks the capacity to impose perspective geometry constrains on $\{c^i\}_{i=0}^7$, which constructs a projective cuboid in the image plane. We thus propose a 3D geometric vanishing point regularization loss, which forces $\{c^i\}_{i=0}^7$ to satisfy perspective geometry during regression, as such the predicted vertices don't require camera calibration data or post preprocessing for refinement~\cite{dwibedi2016deep}. In projective geometry, the two-dimensional perspective projections of mutually parallel lines in three-dimensional space appear to converge at the vanishing point. The required condition for convergence of three lines is that the determinant of the coefficient matrix is zero. The proposed vanishing point loss encodes this geometry constraint (as shown in Figure \ref{fig:vploss}) by minimizing the determinants of all sets of three parallel edges of the vehicle. Formally, taking three parallel lines ${\bf{l}}_{c^0 c^3}$, ${\bf{l}}_{c^1 c^2}$, ${\bf{l}}_{c^5 c^6}$ in {\textbf{F}} as examples (as shown in Figure \ref{fig:vploss}), the vanishing point loss and the coefficient matrix are expressed as:

\begin{flalign}
L_{\text{vp}_{\textbf{F}_1}} = (D_{\text{vp}_{\textbf{F}_1}})^2,
D_{\text{vp}_{\textbf{F}_1}} = \begin{vmatrix}
m_{c^0 c^3} & n_{c^0 c^3} & l_{c^0 c^3} \\
m_{c^1 c^2} & n_{c^1 c^2} & l_{c^1 c^2} \\
m_{c^5 c^6} & n_{c^5 c^6} & l_{c^5 c^6}
\end{vmatrix},
\end{flalign}

where $m_{c^ic^j}x + n_{c^ic^j}y + l_{c^ic^j} = 0$ is the line equation of ${\textbf{l}}_{c^ic^j}$, and $D$ is the determinant of the matrix. $L_{\text{vp}_{\textbf{F}_1}}$ is the first part of $L_{\text{vp}_{\textbf{F}}}$ using the first three lines (${\textbf{l}}_{c^0c^3}$, ${\textbf{l}}_{c^1c^2}$, and ${\textbf{l}}_{c^5c^6}$; see Figure \ref{fig:vploss} for details.). Similarly, we build the second part, $L_{\text{vp}_{\textbf{F}_2}}$, using the last three lines (${\textbf{l}}_{c^0c^3}$, ${\textbf{l}}_{c^4c^7}$, and ${\textbf{l}}_{c^5c^6}$) in the diagonal to form up the final vanishing point regularization $L_{\text{vp}_{\textbf{F}}} = L_{\text{vp}_{\textbf{F}_1}} \! + \! L_{\text{vp}_{\textbf{F}_2}}$ for the \textbf{F} direction, and repeat for the R and S directions. Therefore, the vanishing point loss of the whole vehicle, $ L_{\text{vp}} = L_{\text{vp}_{\textbf{R}}} + L_{\text{vp}_{\textbf{S}}} + L_{\text{vp}_{\textbf{F}}}$.
\subsubsection{3D Perspective Feature}
\begin{figure}[t]
	\centering
	\includegraphics[width=1\linewidth]{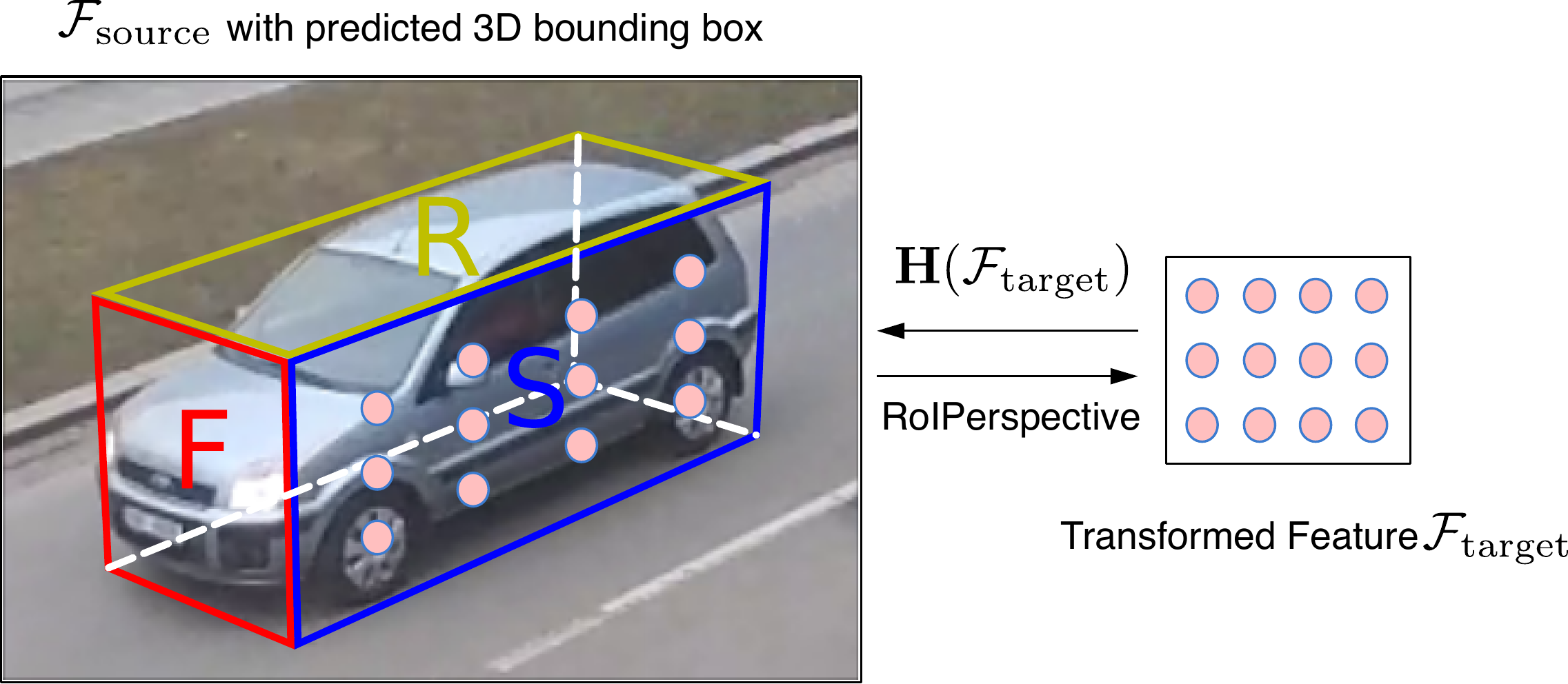}
	\caption{The process of extracting perspective corrected features from {\textbf{S}} using perspective RoI. $\mathcal{F}_{\text{source}}$ and $\mathcal{F}_{\text{target}}$ are the source and target feature maps respectively. To improve visualization, we show the input image overlayed with $\mathcal{F}_{\textbf{S}}$ to show where the predicted 3D bounding box and sampling points (colored by pink) are.}
	\label{fig:carhomopooling}
\end{figure}

Up to this point, the 3D bounding box has already been obtained. To eliminate the viewpoint variance of the 3D bounding box, 3D perspective features are generated in feature-map level by warping each side of the 3D bounding box onto a canonical plane. Since each side of the 3D bounding box is a quadrilateral generated by camera projection, warping them using homography is geometrically correct and a natural choice. In this paper, we adapt the RoI perspective \cite{sun2018textnet} to extract fixed-size feature by mapping each side to the canonical plane. Specifically, suppose that we have a source feature map $\mathcal{F}_{\text{source}}$, which is extracted from the input image using a standard CNN, and a corresponding vehicle side, $Q$. We aim to generate 3D perspective features by mapping the feature inside $Q$ of $\mathcal{F}_{\text{source}}$ to a fixed-size target feature map, $\mathcal{F}_{\text{target}}$. Extracting fixed-size feature from a given region has already been well studied  We first use a four-correspondence DLT \cite{hartley2003multiple} to obtain the homography $\textbf{H}$ between $Q$ and $\mathcal{F}_{\text{target}}$:
\begin{equation}
\left[ {\begin{array}{*{20}{c}}
	q_{x}^i\\
	q_{y}^i\\
	1
	\end{array}} \right] \sim \left[ {\begin{array}{*{20}{c}}
	{\bf{H}}_{11}&{\bf{H}}_{12}&{\bf{H}}_{13}\\
	{\bf{H}}_{21}&{\bf{H}}_{22}&{\bf{H}}_{23}\\
	{\bf{H}}_{31}&{\bf{H}}_{32}& 1
	\end{array}} \right]\left[ {\begin{array}{*{20}{c}}
	t_{x}^i\\
	t_{y}^i\\
	1
	\end{array}} \right],
\label{eqn:homo}
\end{equation}
where $\{t^i\}_{i=1}^4$ and $\{q^i\}_{i=1}^4$ are the four corners of $\mathcal{F}_{\text{target}}$ and $Q$ respectively. Thus given the coordinate of each pixel in $\mathcal{F}_{\text{target}}$, we can obtain their corresponding sampling point in $\mathcal{F}_{\text{source}}$ using $\bf{H}$. In the feature sampling step, the exact value of each sampling point at $\mathcal{F}_{\text{source}}$ can be computed easily using bilinear interpolation at four regularly sampled locations. The sampling step details can be found in \cite{ren2015faster}. Up to this point, the feature inside $Q$ is transformed as a fixed-size target feature map $\mathcal{F}_{\text{target}}$. Figure~\ref{fig:carhomopooling} visualizes the process of generating 3D features from $\mathcal{F}_{\textbf{S}}$. $\mathcal{F}_{\textbf{R}}$ and $\mathcal{F}_{\textbf{F}}$ can be obtained in a similar manner.
\subsection{Feature Fusion Network (FFN)}
\label{sec:fa}
\begin{figure}[!t]
	\includegraphics[width=1\linewidth]{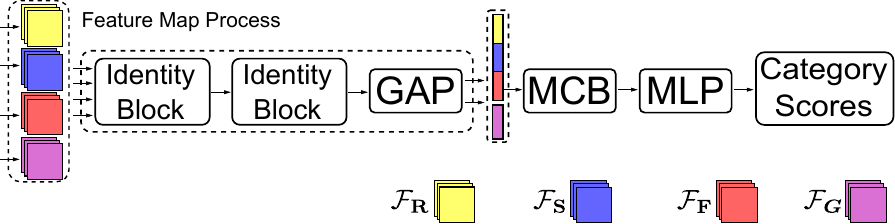}
	\caption{Feature Fusion Network (FFN) architecture.}
	\label{fig:ffn}
\end{figure}

\noindent
Figure \ref{fig:ffn} visualizes the architecture of the FFN, which is designed to merge feature maps extracted from the GN and 3DPN to recognize a given vehicle. Three 3D feature representations  $\mathcal{F}_{\textbf{S}}$, $\mathcal{F}_{\textbf{R}}$, $\mathcal{F}_{\textbf{F}}$ and one global feature $\mathcal{F}^{B}_G$ are processed through two identity blocks \cite{he2016deep}, followed by a global average pooling (GAP) layer, to generate refined feature vectors respectively. Please note that the three feature vectors from {\textbf{F}}, {\textbf{R}}, and {\textbf{S}} are concatenated together to form a single perspective feature vector carrying discriminative perspective information representing different vehicle views. The final feature vector, whose size is 16000, is obtained by applying multi-modal compact bilinear (MCB)~\cite{fukui2016multimodal} pooling on the global and perspective feature vector. The reason for using MCB is that it is normally used to facilitate the joint optimization of two networks generating features which lie on different manifolds.
The two feature vectors are obtained from two different networks (GN vs. 3DPN), i.e., they lie on different manifolds. The final feature vector is passed through two fully-connected (fc) layers of size 2048 and the number of categories, respectively. Up to this point, our full model, which is composed of three network components, can be trained jointly with a single optimization process using the following multi-task loss function:
\begin{equation}
L = \lambda_1 L_{\text{2DGN}} + \lambda_2 L_{\text{3DBranch}} + \lambda_3 L_{\text{CrossEntropy}},
\end{equation}
where $L_{\text{2DGN}}$ is the focal loss \cite{lin2018focal} used to train the GN, $L_{\text{3DBranch}}$ is defined in Equation \ref{eqn:bb3d}, and $L_{\text{CrossEntropy}}$ is the cross entropy loss to train the last softmax layer in the FFN.
\section{Experiments}
\subsection{Implementation and Training Details}
The RetinaNet \cite{lin2018focal} backbone used in the GN is built on MobileNetV2 \cite{sandler2018mobilenetv2}. ResNet101 \cite{he2016deep} with the first 4 stages is selected as the 3DPN architecture. We select different backbones for 3DPN and GN because features produced from them lie on different manifolds. This can enhance the representation of the unified network. In this paper, we adopt a pragmatic 2-step training approach to optimize the whole network. In the first step, we train the GN solely so that it can output the 2D bounding box correctly, which is important to train the 3DPN which takes the 2D bounding box as input. In the second step, we train all three network components, the GN, 3DPN, and FFN, together in an end-to-end manner. $\lambda_1$, $\lambda_2$, $\lambda_3$ are set to 1, 0.1, and 1 respectively. SGD is chosen as our optimizer and its momentum is set to 0.9. The initial learning rate is 0.02, and is divided by 10 after every 15 epochs. The batch size is set to 30. The model optimisation is ceases when training reaches 45 epochs. Each batch takes approximately 2s on a NVIDIA Tesla P100 GPU and in total the model takes about 12 hours to converge.
\subsection{Dataset}
To our best knowledge, the BoxCars dataset \cite{sochor2016boxcars} is only dataset which provides both 3D and 2D bounding box annotations for vehicle recognition in the computer vision community. Therefore, we use it to evaluate our model. BoxCars contains 63,750 images, which are collected from 21,250 vehicles of 27 different makes. All images are taken from surveillance cameras. BoxCars consists of two challenging tasks: classification and verification. Regarding the classification task, the dataset is split into two subsets: {\textit{Medium}} and {\textit{Hard}}. The {\textit{Hard} protocol has 87 categories and contains 37,689 training images and 18,939 testing images. The {\textit{Medium}} protocol is composed of 77 categories and has 40,152 and 19,590 images for training and testing respectively. The main difference between the {\textit{Medium}} and {\textit{Hard}} splits is that {\textit{Hard}} considers make, model, submodel, and model year; while {\textit{Medium}} does not differentiate model year. With respect to the verification task, BoxCars has three well defined protocols that provide {\textit{Easy}}, {\textit{Medium}}, and {\textit{Hard}} cases. The {\textit{Easy}} protocol is composed of pairs of vehicle images recorded from the same unseen camera. Camera identities are no longer fixed in the {\textit{Medium}} protocol. The {\textit{Hard}} protocol not only draws vehicle pairs from different unseen cameras, but also takes into account vehicle model years.
\subsection{Vehicle Classification Results}
\begin{table*}[t]
	\caption{Overall classification accuracy on BoxCars dataset. {\textit{M}} and {\textit{H}} represent the {\textit{Medium}} and {\textit{Hard}} splits. Top-1 and -5 accuracy are denoted as T-1 and T-5.}
	\begin{center}
		
		\begin{tabular}{l|c|c|c|c|c|c|c|c} 
			\hline			
			\textbf{Method}  & Input Size & Detection? & 3D? & Attention? & {\textit{M}} T-1 & {\textit{M}} T-5 &  {\textit{H}} T-1 &  {\textit{H}} T-5\\
			
			\specialrule{0.12em}{0em}{0em}
			Faster-RCNN   &$256 \times 256$ & \ding{51}& \ding{55}&  \ding{55}& 67.23  & -  & 62.73  & - \\
			2DGN-det (RetinaNet)   &$256 \times 256$&\ding{51} &  \ding{55}   & \ding{55} & 66.52 & -  & 59.4  & -\\	
			Ours-{\textit{det}}   &$256 \times 256$&    \ding{51} & \ding{51}   & \ding{51}  & {\textbf{78.45}} & {\textbf{93.39}}  & {\bf{75.18}}  & {\textbf{91.53}} \\
			\specialrule{0.12em}{0em}{0em}

			NTS  &$224 \times 224$&\ding{55}  &\ding{55} &\ding{51}& 80.40& 92.37  & 76.31  & 90.42\\
			
			DFL &$224 \times 224$&\ding{55}  &\ding{55} &\ding{51}& 76.78& 91.94  & 70.25 & 88.405\\
			BoxCar   &$224 \times 224$&   \ding{55}  & \ding{55} & \ding{51}& 75.4  &  90.1 &  73.1 & 89\\
			
			RACNN   &$224 \times 224$&\ding{55}  &\ding{55} &\ding{51}& 72.21& 88.47  & 67.5 & 86.83\\
			STN  &$224 \times 224$&\ding{55}  &\ding{55} &\ding{51}& 64.33& 81.92  & 59.76 & 80.13\\
			
			3DPN-{\textit{cls}}   &$224 \times 224$&    \ding{55} & \ding{51}   & \ding{51}  & 80.31 & 92.04  & 76.68  &  90.71 \\
			Ours-{\textit{cls}} &$224 \times 224$& \ding{55} &\ding{51} &\ding{51}& {\textbf{81.27}} & {\textbf{93.82}}  & {\bf{77.08}}  & {\textbf{91.97}} \\
			\hline		
		\end{tabular}
	\end{center}
	\label{tab:clsbaseline}
\end{table*}
\begin{figure*}[t]
	\centering
	\begin{subfigure}{0.62\linewidth}
		\centering
		\includegraphics[width=0.19\linewidth]{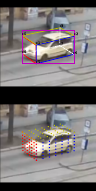}
		\includegraphics[width=0.19\linewidth]{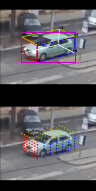}
		\includegraphics[width=0.19\linewidth]{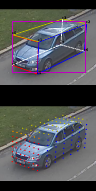}
		\includegraphics[width=0.19\linewidth]{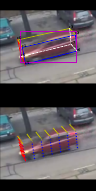}
		\includegraphics[width=0.19\linewidth]{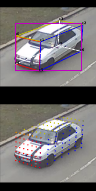}
		\caption{\label{fig:goodclsresults}The correctly predicted examples.}
	\end{subfigure}
	\begin{subfigure}{0.37\linewidth}
		\centering
		\includegraphics[width=0.32\linewidth]{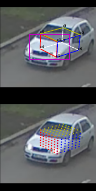}
		\includegraphics[width=0.32\linewidth]{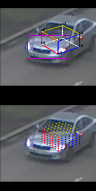}
		\includegraphics[width=0.32\linewidth]{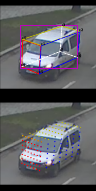}
		\caption{\label{fig:badclsresults}The mispredicted examples.}
	\end{subfigure}
	\caption{Qualitative results visualization of Ours-{\textit{det}} on BoxCars dataset. (\subref{fig:goodclsresults}): examples in which the 2D and 3D bounding box are correctly predicted. (\subref{fig:badclsresults}): examples containing errors in prediction.}
	\label{fig:moreresults}
\end{figure*}
\subsubsection{Baselines}
Since our model recognize vehicles from a single image, single-image based methods, including BoxCar \cite{sochor2016boxcars}, Faster-RCNN \cite{ren2015faster}, RetinaNet \cite{lin2018focal}, NTS \cite{yang2018learning}, DFL \cite{wang2018learning}, RACNN \cite{fu2017look}, STN \cite{jaderberg2015spatial}, are selected to compare with our method. These methods are divided into two evaluation categories: (1) detection-like (\textit{det}-like) networks (2DGN-det, FasterRCNN), in which localization and classification of the vehicle are performed simultaneously; and (2) classification-like (\textit{cls}-like) networks (NTS, DFL, RACNN, and STN) in which vehicles are cropped using annotated bounding box before network training. With respect to classification-like networks, all images are resized to $224 \times 224$. Regarding detection-like networks, images are resized to the same scale of 256 pixels as in \cite{lin2017feature}. To make fair comparison, we use the official implementations of these methods without any parameter changes.
\subsubsection{\textit{det}-like network results}
The upper half of Table \ref{tab:clsbaseline} shows the results of {\textit{det}}-like networks. One can see that Ours-{\textit{det}} surpasses all {\textit{det}}-like baselines by a significant margin.
Since Faster-RCNN shares the same backbone with the GN in Ours-{\textit{det}} and 2DGN-{\textit{det}} is the detection part of Ours-{\textit{det}}, we confirm that the additional 3D feature representation significantly improves the performance obtained compared to using traditional 2D features. From Table \ref{tab:clsbaseline}, one can see that 2DGN-{\textit{det}} and Faster-RCNN do not have a top-5 accuracy recorded. This is because 2DGN-{\textit{det}} and Faster-RCNN output confidence scores of predicted boxes. After non-maximum suppression, the boxes with high confidence scores are merged, and as such there is only a top-1 accuracy. Although non-maximum suppression is also performed in our method, we can still obtain top-5 accuracy due to the use of the softmax layer in the FFN.
\subsubsection{\textit{cls}-like network results}
To make a comparison between \textit{cls}-like baselines and the proposed approach, we modify the 2D processing component of our model. Specifically, MobileNetV2-based RetinaNet is replaced with a vanilla MobileNetV2, in which the last global average pooling layer and following classification layer are removed. Therefore the output of this network is used as a global feature for the vehicle. The modified model for \textit{cls}-like experiments is denoted Ours-{\textit{cls}}. In addition, we separate 3DPN-{\textit{cls}} to do isolated component testing. Since 3DPN is not able to produce 2D bounding box, we feed ground truth 2D bounding box to 3DPN-{\textit{cls}} to adapt it as a classification network. Please note that 3DPN do not have {\textit{det}} version simply because that it cannot produce 2D bounding box.

The second half of Table \ref{tab:clsbaseline} showcases overall classification accuracy (percent) for \textit{cls}-like networks. We observe that Ours-{\textit{cls}} consistently performs better than all baseline models with respect to classification accuracy in both the {\textit{Medium}} and {\textit{Hard}} splits. One can see that STN and RACNN perform poorly among all \textit{cls}-like methods, as they only search for the most discriminative part of a vehicle. This strategy discards parts of the global feature, which captures important pose information and other subtle details. Moreover, an affine transformation used in STN significantly increases the difficulty of vehicle viewpoint normalization. This is because an affine transformation of the 2D vehicle bounding box distorts the shape of the vehicle, and does not consider its 3D geometry. We next compare Ours-\textit{cls} with NTS, DFL, and BoxCar baselines, which extract discriminative features without considering the 3D geometry. From the results, we conjecture that the combined 2D and 3D representation used in our method has better a capability for distinguishing vehicle details than other methods. It is worthy note that 3DPN-\textit{cls} can already achieve comparable performance to previous state-of-the-art works.
\subsubsection{Qualitative results}
Figure \ref{fig:moreresults} visualizes qualitative results on BoxCars images. 
In Figure \ref{fig:moreresults} (\subref{fig:goodclsresults}), we see that Ours-{\textit{det}} is able to determine the correct 2D location and 3D bounding box estimation of the vehicle. Figure \ref{fig:moreresults} (\subref{fig:badclsresults}) shows some mis-estimated images.
The first two columns show 3D bounding box regression performance, in situations where the 2D bounidng box is incorrect, and the 3D estimation cannot recover from the earlier error. The last column shows a case where 2D location is predicted correctly and the 3D box estimator fails. We see that the 3D bounding box estimation tries to compensate for errors made by the 2D bounding box estimation. In addition, the sampling points on {\textbf{F}}, {\textbf{R}}, and {\textbf{S}} are also shown. One can see that the sampling points perfectly cover the three main sides of the vehicles, and therefore extract perspective invariant features.
\subsection{Ablation experiments}
An ablation study is conducted to analyze the effectiveness of the individual proposed components including 3D perspective feature (3DPF) and the VP regularization loss in the 3D bounding box regression component.
\subsubsection{3D Perspective Feature vs. Attention-Based Feature}
In this experiment we compare the performance of the 3D perspective feature to the attention-based feature, which is frequently used by previous works, on the {\textit{Hard}} and {\textit{Medium}} splits. The perspective RoI layers in {Ours-{\textit{det}}} are replaced with RoI Align \cite{ren2015faster} to simulate the attention-based feature (ABF) obtained by attention mechanism. Specifically, three main sides of a vehicle are located using 2D rectangle bounding boxes in ABF rather than geometrically correct quadrilaterals in 3DPF. The results are shown in Table \ref{tab:ablationroivs}. It can be observed that 3DPF achieves approximately 6.1\% and 2.1\% improvement on {\textit{Medium}} and {\textit{Hard}} splits.
\begin{figure*}[!t]
	\centering
	\includegraphics[width=0.33\linewidth]{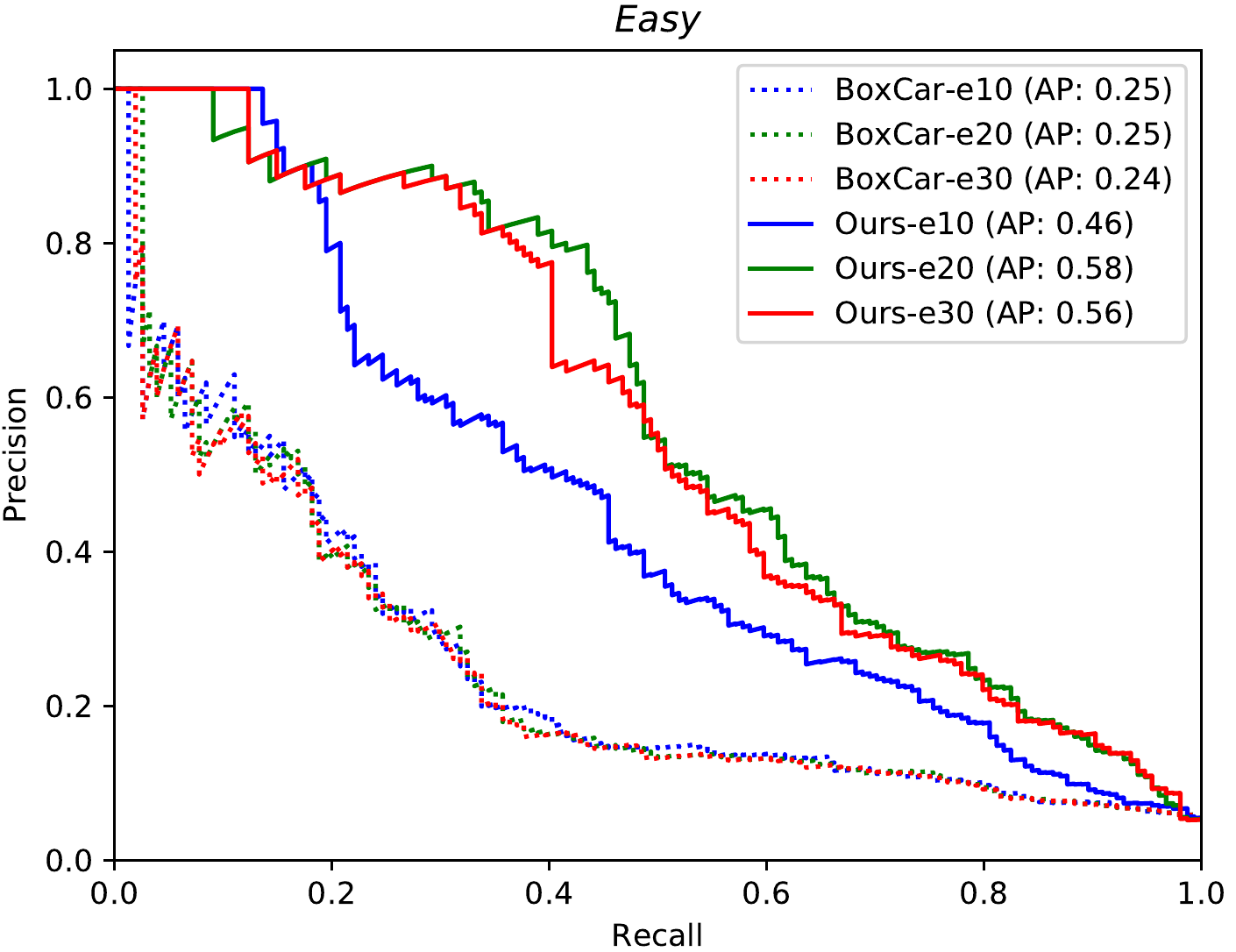}
	\includegraphics[width=0.33\linewidth]{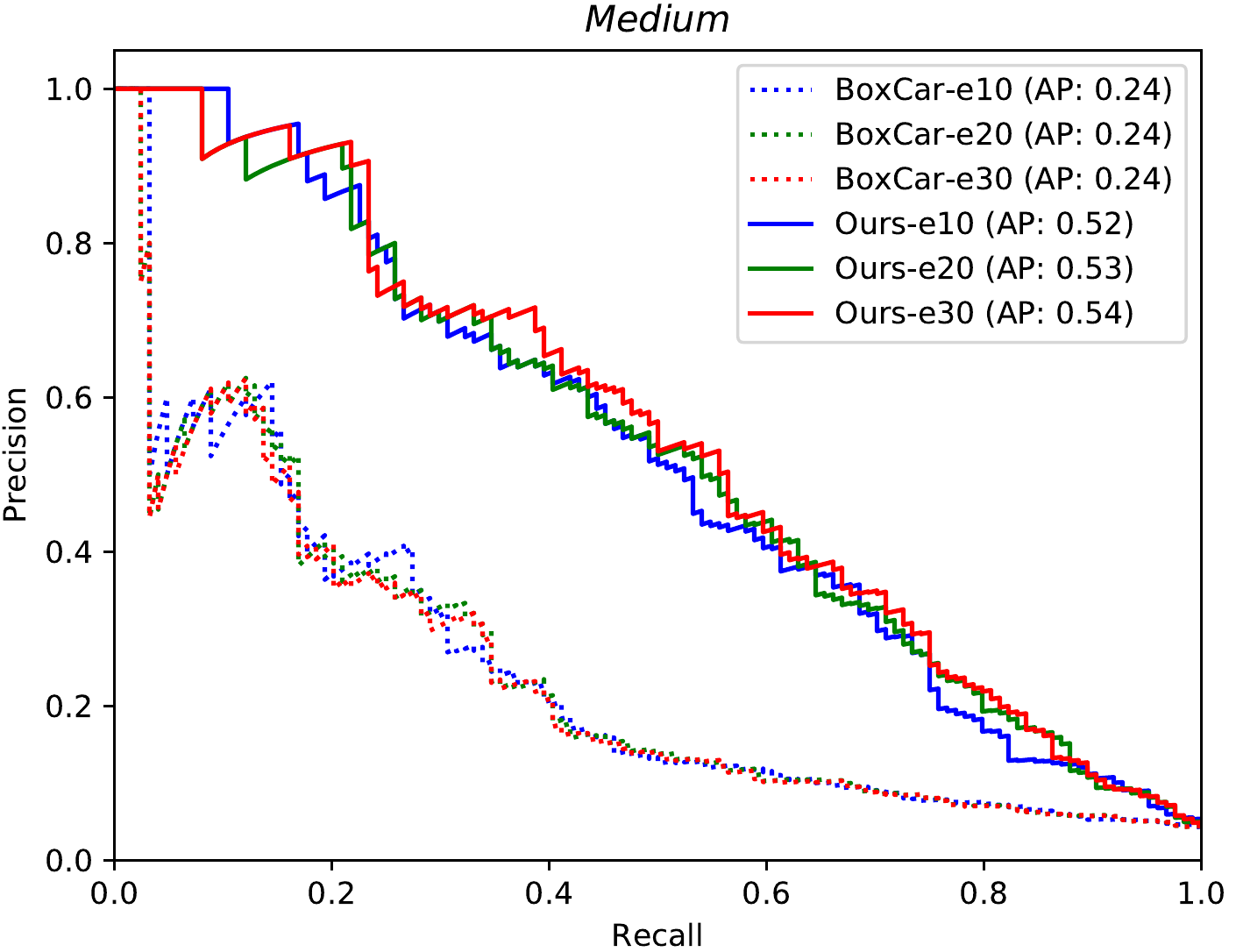}
	\includegraphics[width=0.33\linewidth]{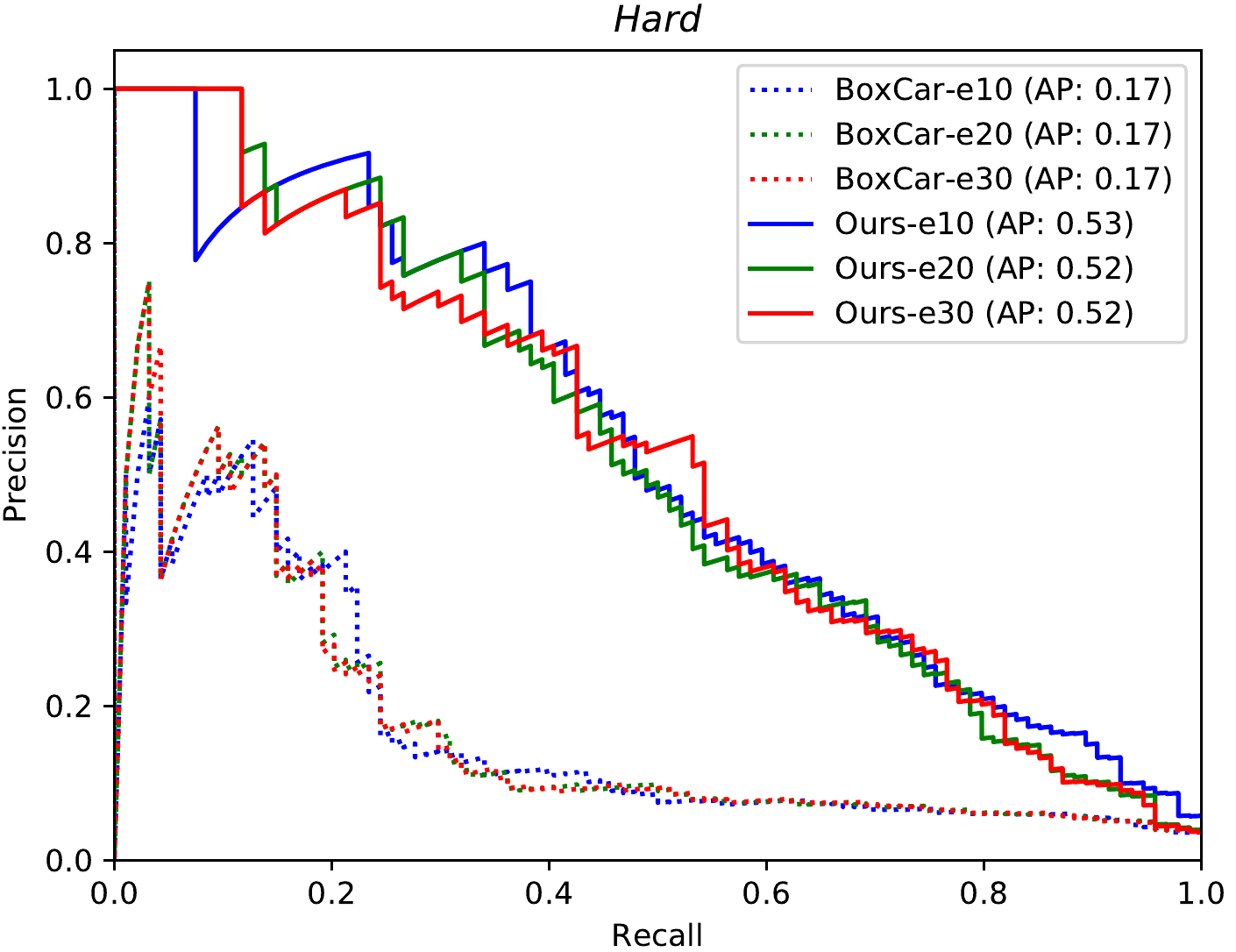}
	\caption{Precision-Recall (PR) curves of different models with different training epoch (e denotes training epoch x). Three verification protocols {\textit{Easy}}, {\textit{Medium}}, and {\textit{Hard}} are shown in the figure. Average Precision (AP) is given in the plot legends. Baseline results (shown as BoxCar-ex) are taken from \cite{sochor2016boxcars}.}
	\label{fig:verfications}
\end{figure*}
\begin{table}
	\centering
	\caption{3DPF vs. ABF: Classification accuracy results with different kinds of feature types.}
	\begin{tabular}{c|c|c|c|c}
		\hline					
		Feature Type       &  {\textit{M.}} T-1 & {\textit{M.}} T-5       &   {\textit{H.}} T-1 & {\textit{H.}} T-5  \\
		\specialrule{0.12em}{0em}{0em}  
		ABF    &73.97& 91.23 & {73.65} & {91.13} \\ 
		3DPF   & {\textbf{78.45}} & {\textbf{93.39}} & {\textbf{75.18}}   & {\textbf{91.53}} \\
		\hline					
	\end{tabular}
	\label{tab:ablationroivs}
\end{table}
\begin{table}
	\small
	\centering
	\setlength\tabcolsep{4pt}
	\caption{Evaluation of 3D bounding box localization and quality using the percentage of correct keypoints (PCK) and the proposed cuboid quality (CQ). $e$ stands for the number of training epochs.}
	\begin{tabular}{l|cc|cc|cc}
		\hline			
		& \multicolumn{2}{c|}{e=5}   & \multicolumn{2}{c|}{e=10} &\multicolumn{2}{c}{e=15}  \\
		& PCK & CQ  & PCK & CQ & PCK & CQ   \\
		\specialrule{0.12em}{0em}{0em}  
		Ours-{\textit{det}}    & \textbf{85.35}& 1.48 &  \textbf{85.66} &  \textbf{1.98} & \textbf{87.15} & \textbf{2.12} \\ 
		Ours-{\textit{det}} w/0 vp loss   & 85.03 & 1.48 & 85.58  & 1.64 & 86.70 & 1.69\\
		\hline					
	\end{tabular}
	\label{tab:ablationvptab}
\end{table}
\begin{table}[t]
	\centering
	\caption{Vanishing point loss: A geometrically correct 3D bounding box gives gain to the proposed method.}
	\begin{tabular}{l|c|c|c|c}
		\hline					
		Feature Type       &  {\textit{M.}} T-1 & {\textit{M.}} T-5       &   {\textit{H.}} T-1 & {\textit{H.}} T-5  \\
		\specialrule{0.12em}{0em}{0em}  
		Ours-{\textit{det}}  & {\textbf{78.45}} & {\textbf{93.39}}  & {\bf{75.18}}  & {\textbf{91.53}} \\
		Ours-{\textit{det}} w/o vp loss   & 78.32 & 93.33& 74.71& 89.94 \\
		
		\specialrule{0.12em}{0em}{0em}
		Ours-{\textit{cls}} & {\textbf{81.27}} & {\textbf{93.82}}  & {\bf{77.08}}  & {\textbf{91.97}}\\
		Ours-{\textit{cls}} w/o vp loss   & 80.92 & 92.88& 76.84& 90.50\\
		\hline					
	\end{tabular}
	\label{tab:ablationvplosscls}
\end{table}
\subsubsection{VP regularization}
We evaluate the proposed VP regularization loss on 3D bounding box detection.  
Table \ref{tab:ablationvptab} reports the results obtained in the {\textit{Hard}} split in terms of two metrics, the percentage of correct points (PCK) \cite{tulsiani2015viewpoints} and the proposed cube quality (CQ). A predicted vertex is considered to be correct if it lies within $0.1 \times \text(max(height,width))$ pixels of the ground truth annotation of the vertex.  CQ is computed via $-\log{L_{\text{VP}}}$. From the results, we can see that the 3D bounding box obtained from our method with VP regularization consistently outperforms that of \cite{dwibedi2016deep} in terms of both metrics. Apart from this, we also test the proposed method without using vp loss in terms of the classification and detection tasks. The results can be found in Table \ref{tab:ablationvplosscls}.
\subsection{Vehicle Verification}
Vehicle verification is the problem of determining whether two gallery samples belong to the same category. It is an important and challenging task for intelligence transportation, especially when the system is working under new scenarios with unseen and misaligned categories.

To demonstrate the generality and robustness of the proposed method, we conduct experiments on the verification task of BoxCars. In this experiment, we follow the same method of \cite{sochor2016boxcars} to perform verification, i.e., 3,000 image pairs are randomly selected to test the performance of various models in each case.

For all networks, we use the output of the second last layer (the layer preceding the last softmax classification layer) as the representation feature vector for the given image. For each image pair, we use the cosine distance \cite{taigman2014deepface} to obtain the similarity of two gallery images, which is then used to compute precision, recall, and average precision.

The precision-recall (PR) curves presented in Figure~\ref{fig:verfications} show that the proposed approach outperforms the baseline method~\cite{sochor2016boxcars} on all three dataset protocols. The performance gain of our method provides an absolute performance gain of 33\% in Average Precision (AP) on {\textit{Easy}}, and an even better 36\% AP on the {\textit{Hard}} split. It is worth noting that the size of feature vector of~\cite{sochor2016boxcars} is 4096 while ours is 2048, which indicates a better data distribution and faster speed for model inference.
\section{Conclusions}
In this paper, we propose a unified framework to perform vehicle classification, which takes full advantage of both the 2D and 3D perspective representations. The proposed method achieves the state-of-the-art results both in car classification and verification in the BoxCars dataset. Furthermore, we propose vanishing point regularization for cuboid detection, which is intuitively appealing and geometrically explainable, and avoids the need for a post detection refinement processes, as used by existing methods. Last but not least, the proposed 3DPF is able to extract features correctly from 3D bounding boxes which warped by perspective transformation.
\bibliographystyle{aaai}
\bibliography{egbib}

\begin{thebibliography}{}

\bibitem[\protect\citeauthoryear{Dwibedi \bgroup et al\mbox.\egroup
  }{2016}]{dwibedi2016deep}
Dwibedi, D.; Malisiewicz, T.; Badrinarayanan, V.; and Rabinovich, A.
\newblock 2016.
\newblock Deep cuboid detection: Beyond 2d bounding boxes.
\newblock {\em arXiv preprint arXiv:1611.10010}.

\bibitem[\protect\citeauthoryear{Fu, Zheng, and Mei}{2017}]{fu2017look}
Fu, J.; Zheng, H.; and Mei, T.
\newblock 2017.
\newblock Look closer to see better: Recurrent attention convolutional neural
  network for fine-grained image recognition.
\newblock In {\em Proceedings of the IEEE conference on computer vision and
  pattern recognition},  4438--4446.

\bibitem[\protect\citeauthoryear{Fukui \bgroup et al\mbox.\egroup
  }{2016}]{fukui2016multimodal}
Fukui, A.; Park, D.~H.; Yang, D.; Rohrbach, A.; Darrell, T.; and Rohrbach, M.
\newblock 2016.
\newblock Multimodal compact bilinear pooling for visual question answering and
  visual grounding.
\newblock {\em arXiv preprint arXiv:1606.01847}.

\bibitem[\protect\citeauthoryear{Gupta \bgroup et al\mbox.\egroup
  }{2011}]{gupta20113d}
Gupta, A.; Satkin, S.; Efros, A.~A.; and Hebert, M.
\newblock 2011.
\newblock From 3d scene geometry to human workspace.
\newblock In {\em Computer Vision and Pattern Recognition (CVPR), 2011 IEEE
  Conference on},  1961--1968.
\newblock IEEE.

\bibitem[\protect\citeauthoryear{Hartley and
  Zisserman}{2003}]{hartley2003multiple}
Hartley, R., and Zisserman, A.
\newblock 2003.
\newblock {\em Multiple view geometry in computer vision}.
\newblock Cambridge university press.

\bibitem[\protect\citeauthoryear{He \bgroup et al\mbox.\egroup
  }{2016}]{he2016deep}
He, K.; Zhang, X.; Ren, S.; and Sun, J.
\newblock 2016.
\newblock Deep residual learning for image recognition.
\newblock In {\em Proceedings of the IEEE conference on computer vision and
  pattern recognition},  770--778.

\bibitem[\protect\citeauthoryear{He \bgroup et al\mbox.\egroup
  }{2017}]{he2017mask}
He, K.; Gkioxari, G.; Doll{\'a}r, P.; and Girshick, R.
\newblock 2017.
\newblock Mask r-cnn.
\newblock In {\em Computer Vision (ICCV), 2017 IEEE International Conference
  on},  2980--2988.
\newblock IEEE.

\bibitem[\protect\citeauthoryear{He, Shao, and Tan}{2015}]{he2015recognition}
He, H.; Shao, Z.; and Tan, J.
\newblock 2015.
\newblock Recognition of car makes and models from a single traffic-camera
  image.
\newblock {\em IEEE Transactions on Intelligent Transportation Systems}
  16(6):3182--3192.

\bibitem[\protect\citeauthoryear{Hedau, Hoiem, and
  Forsyth}{2012}]{hedau2012recovering}
Hedau, V.; Hoiem, D.; and Forsyth, D.
\newblock 2012.
\newblock Recovering free space of indoor scenes from a single image.
\newblock In {\em 2012 IEEE Conference on Computer Vision and Pattern
  Recognition},  2807--2814.
\newblock IEEE.

\bibitem[\protect\citeauthoryear{Jaderberg \bgroup et al\mbox.\egroup
  }{2015}]{jaderberg2015spatial}
Jaderberg, M.; Simonyan, K.; Zisserman, A.; et~al.
\newblock 2015.
\newblock Spatial transformer networks.
\newblock In {\em Advances in neural information processing systems},
  2017--2025.

\bibitem[\protect\citeauthoryear{Krause \bgroup et al\mbox.\egroup
  }{2013}]{krause20133d}
Krause, J.; Stark, M.; Deng, J.; and Fei-Fei, L.
\newblock 2013.
\newblock 3d object representations for fine-grained categorization.
\newblock In {\em Proceedings of the IEEE International Conference on Computer
  Vision Workshops},  554--561.

\bibitem[\protect\citeauthoryear{Krause \bgroup et al\mbox.\egroup
  }{2014}]{krause2014learning}
Krause, J.; Gebru, T.; Deng, J.; Li, L.; and Fei-Fei, L.
\newblock 2014.
\newblock Learning features and parts for fine-grained recognition.
\newblock In {\em 2014 22nd International Conference on Pattern Recognition},
  26--33.

\bibitem[\protect\citeauthoryear{Lin \bgroup et al\mbox.\egroup
  }{2014}]{lin2014jointly}
Lin, Y.-L.; Morariu, V.~I.; Hsu, W.; and Davis, L.~S.
\newblock 2014.
\newblock Jointly optimizing 3d model fitting and fine-grained classification.
\newblock In {\em European Conference on Computer Vision},  466--480.
\newblock Springer.

\bibitem[\protect\citeauthoryear{Lin \bgroup et al\mbox.\egroup
  }{2017}]{lin2017feature}
Lin, T.-Y.; Doll{\'a}r, P.; Girshick, R.~B.; He, K.; Hariharan, B.; and
  Belongie, S.~J.
\newblock 2017.
\newblock Feature pyramid networks for object detection.
\newblock In {\em CVPR}, volume~1, ~4.

\bibitem[\protect\citeauthoryear{Lin \bgroup et al\mbox.\egroup
  }{2018}]{lin2018focal}
Lin, T.-Y.; Goyal, P.; Girshick, R.; He, K.; and Doll{\'a}r, P.
\newblock 2018.
\newblock Focal loss for dense object detection.
\newblock {\em IEEE transactions on pattern analysis and machine intelligence}.

\bibitem[\protect\citeauthoryear{Manhardt, Kehl, and
  Gaidon}{2019}]{manhardt2019roi}
Manhardt, F.; Kehl, W.; and Gaidon, A.
\newblock 2019.
\newblock Roi-10d: Monocular lifting of 2d detection to 6d pose and metric
  shape.
\newblock In {\em Proceedings of the IEEE Conference on Computer Vision and
  Pattern Recognition},  2069--2078.

\bibitem[\protect\citeauthoryear{Mousavian \bgroup et al\mbox.\egroup
  }{2017}]{mousavian20173d}
Mousavian, A.; Anguelov, D.; Flynn, J.; and Ko{\v{s}}eck{\'a}, J.
\newblock 2017.
\newblock 3d bounding box estimation using deep learning and geometry.
\newblock In {\em Computer Vision and Pattern Recognition (CVPR), 2017 IEEE
  Conference on},  5632--5640.
\newblock IEEE.

\bibitem[\protect\citeauthoryear{Ren \bgroup et al\mbox.\egroup
  }{2015}]{ren2015faster}
Ren, S.; He, K.; Girshick, R.; and Sun, J.
\newblock 2015.
\newblock Faster r-cnn: Towards real-time object detection with region proposal
  networks.
\newblock In {\em Advances in neural information processing systems},  91--99.

\bibitem[\protect\citeauthoryear{Sandler \bgroup et al\mbox.\egroup
  }{2018}]{sandler2018mobilenetv2}
Sandler, M.; Howard, A.; Zhu, M.; Zhmoginov, A.; and Chen, L.-C.
\newblock 2018.
\newblock Mobilenetv2: Inverted residuals and linear bottlenecks.
\newblock In {\em Proceedings of the IEEE Conference on Computer Vision and
  Pattern Recognition},  4510--4520.

\bibitem[\protect\citeauthoryear{Simonelli \bgroup et al\mbox.\egroup
  }{2019}]{simonelli2019disentangling}
Simonelli, A.; Bul{\`o}, S. R.~R.; Porzi, L.; L{\'o}pez-Antequera, M.; and
  Kontschieder, P.
\newblock 2019.
\newblock Disentangling monocular 3d object detection.
\newblock {\em arXiv preprint arXiv:1905.12365}.

\bibitem[\protect\citeauthoryear{Sochor, Herout, and
  Havel}{2016}]{sochor2016boxcars}
Sochor, J.; Herout, A.; and Havel, J.
\newblock 2016.
\newblock Boxcars: 3d boxes as cnn input for improved fine-grained vehicle
  recognition.
\newblock In {\em 2016 IEEE Conference on Computer Vision and Pattern
  Recognition (CVPR)},  3006--3015.

\bibitem[\protect\citeauthoryear{Sun \bgroup et al\mbox.\egroup
  }{2018}]{sun2018textnet}
Sun, Y.; Zhang, C.; Huang, Z.; Liu, J.; Han, J.; and Ding, E.
\newblock 2018.
\newblock Textnet: Irregular text reading from images with an end-to-end
  trainable network.
\newblock In {\em Asian Conference on Computer Vision},  83--99.
\newblock Springer.

\bibitem[\protect\citeauthoryear{Taigman \bgroup et al\mbox.\egroup
  }{2014}]{taigman2014deepface}
Taigman, Y.; Yang, M.; Ranzato, M.; and Wolf, L.
\newblock 2014.
\newblock Deepface: Closing the gap to human-level performance in face
  verification.
\newblock In {\em Proceedings of the IEEE conference on computer vision and
  pattern recognition},  1701--1708.

\bibitem[\protect\citeauthoryear{Tulsiani and
  Malik}{2015}]{tulsiani2015viewpoints}
Tulsiani, S., and Malik, J.
\newblock 2015.
\newblock Viewpoints and keypoints.
\newblock In {\em Proceedings of the IEEE Conference on Computer Vision and
  Pattern Recognition},  1510--1519.

\bibitem[\protect\citeauthoryear{Wang, Morariu, and
  Davis}{2018}]{wang2018learning}
Wang, Y.; Morariu, V.~I.; and Davis, L.~S.
\newblock 2018.
\newblock Learning a discriminative filter bank within a cnn for fine-grained
  recognition.
\newblock In {\em Proceedings of the IEEE Conference on Computer Vision and
  Pattern Recognition},  4148--4157.

\bibitem[\protect\citeauthoryear{Xu, Anguelov, and
  Jain}{2018}]{xu2018pointfusion}
Xu, D.; Anguelov, D.; and Jain, A.
\newblock 2018.
\newblock Pointfusion: Deep sensor fusion for 3d bounding box estimation.
\newblock In {\em Proceedings of the IEEE Conference on Computer Vision and
  Pattern Recognition},  244--253.

\bibitem[\protect\citeauthoryear{Yang \bgroup et al\mbox.\egroup
  }{2018}]{yang2018learning}
Yang, Z.; Luo, T.; Wang, D.; Hu, Z.; Gao, J.; and Wang, L.
\newblock 2018.
\newblock Learning to navigate for fine-grained classification.
\newblock In {\em Proceedings of the European Conference on Computer Vision
  (ECCV)},  420--435.

\end{thebibliography}

\end{document}